\documentclass{article}

\usepackage{arxiv}
\usepackage[utf8]{inputenc} %
\usepackage[T1]{fontenc}    %
\usepackage{hyperref}       %

\usepackage{graphicx}
\usepackage{afterpage}
\usepackage{caption}
\usepackage{subcaption}
\usepackage{xstring}
\usepackage{xcolor}
\usepackage{makecell}
\usepackage{todonotes}

\newcommand{\lm}{LINEMOD}
\newcommand{\lmo}{Occlusion LINEMOD}

\newcommand{\addnosymm}{\textsc{add-$0.1$d}}
\newcommand{\addsymm}{\textsc{add-s-$0.1$d}}
\newcommand{\add}{\textsc{add(-s)-$0.1$d}}
\newcommand{\reproj}{\textsc{reproj-$5$px}}
\newcommand{\reprojs}{\textsc{reproj-s-$5$px}}
\newcommand{\degcm}{\textsc{5cm/$5^\circ$}}
\newcommand{\degcms}{\textsc{5cm/$5^\circ$-s}}

\newcommand{\score}[1]{${\StrSubstitute{#1}{,}{.}}$}
\newcommand{\scorebf}[1]{$\mathbf{\StrSubstitute{#1}{,}{.}}$}

\newcommand{\SO}[1]{ \textnormal{SO}(#1) }

\title{Pose Proposal Critic: Robust Pose Refinement by Learning Reprojection Errors}

\author{
 Lucas Brynte \\
  Chalmers University of Technology\\
  \texttt{brynte@chalmers.se} \\
 \And
 Fredrik Kahl \\
  Chalmers University of Technology\\
  \texttt{fredrik.kahl@chalmers.se} \\
}
\date{}

\def\etal{\emph{et al}.}

\begin{document}

\maketitle

\begin{abstract}
In recent years, considerable progress has been made for the task of rigid object pose estimation from a single RGB-image, but achieving robustness to partial occlusions remains a challenging problem.
Pose refinement via rendering has shown promise in order to achieve improved results, in particular, when data is scarce.

In this paper we focus our attention on pose refinement, and show how to push the state-of-the-art further in the case of partial occlusions. The proposed pose refinement method leverages on a simplified learning task, where a CNN is trained to estimate the reprojection error between an \emph{observed} and a \emph{rendered} image. We experiment by training on purely synthetic data as well as a mixture of synthetic and real data.
Current state-of-the-art results are outperformed for two out of three metrics on the \lmo{} benchmark, while performing on-par for the final metric.

\end{abstract}

\section{Introduction}
Accurately estimating the 3D location and orientation of an object from a single image, a.k.a.\ rigid object pose estimation, has many important real world applications, such as robotic manipulation, augmented reality and autonomous driving.
Although the problem has commonly been addressed by exploiting RGB-D cameras, e.g., \cite{Brachmann_2014_ECCV}, this introduces an increased cost of hardware, sensitivity to sunlight, and unreliable / missing depth measurements for reflective / transparent objects.
In recent years, more attention has been put to RGB-only pose estimation, and although considerable progress has been made, a major challenge remains in achieving robustness to partial occlusions.
To this end, rendering-based pose refinement methods have shown promise in order to achieve improved results, but their full potential remains unexplored.

In this paper we revisit pose refinement via rendering, and focus specifically on how to further improve on the robustness of such methods, in particular with respect to partial occlusions. Our method can hence be used to refine the estimates of any pose algorithm and we will give several experimental demonstrations that this is indeed achieved for different algorithms. Naturally, we will also compare to other refinement methods.

\begin{figure}[t]
    \centering
    \begin{minipage}{0.8\textwidth}

    \newlength{\origheight}
    \newlength{\origwidth}
    \settoheight{\origheight}{\includegraphics{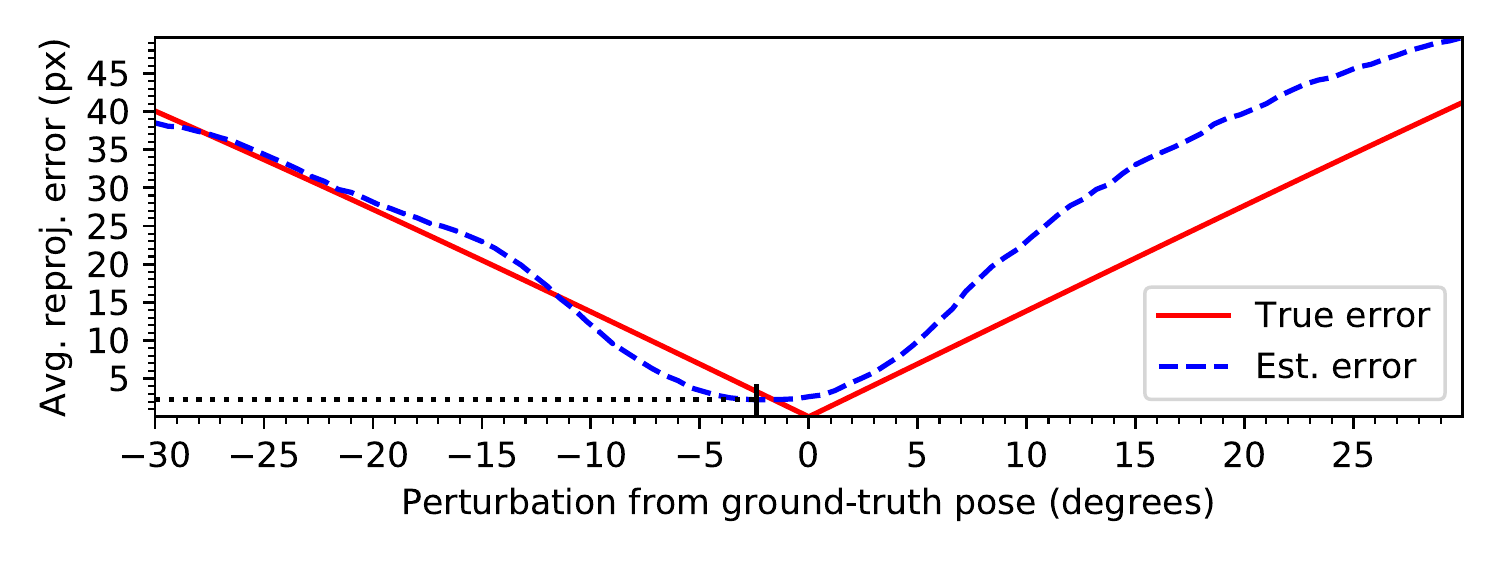}}
    \settowidth{\origwidth}{\includegraphics{figs/cat_errplot_test_occl_f157_rot_xyz_delta_0_17pi.pdf}}
    \newlength{\newheight}
    \setlength\newheight{\textwidth*\ratio{\origheight}{\origheight+\origwidth}}

    \centering
    \subcaptionbox{}[0.900\newheight]{\vspace{0.050\newheight}\includegraphics[width=0.900\newheight]{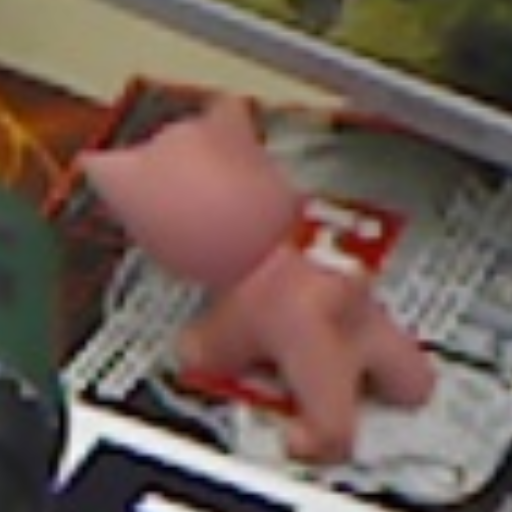}\vspace{0.050\newheight}
    \vspace{-0.150\newheight}
    }
    \hfill
    \subcaptionbox{}[0.99\textwidth-\newheight]{\includegraphics[width=0.99\textwidth-\newheight]{figs/cat_errplot_test_occl_f157_rot_xyz_delta_0_17pi.pdf}
    \vspace{-0.150\newheight}
    }\\

    \subcaptionbox{}[0.900\newheight]{\vspace{0.050\newheight}\includegraphics[width=0.900\newheight]{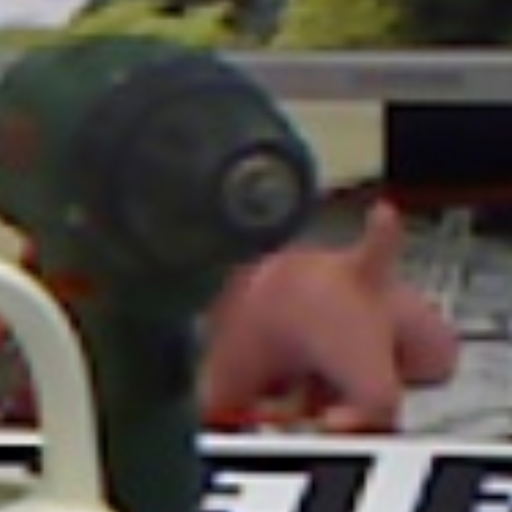}\vspace{0.050\newheight}
    \vspace{-0.150\newheight}
    }
    \hfill
    \subcaptionbox{}[0.99\textwidth-\newheight]{\includegraphics[width=0.99\textwidth-\newheight]{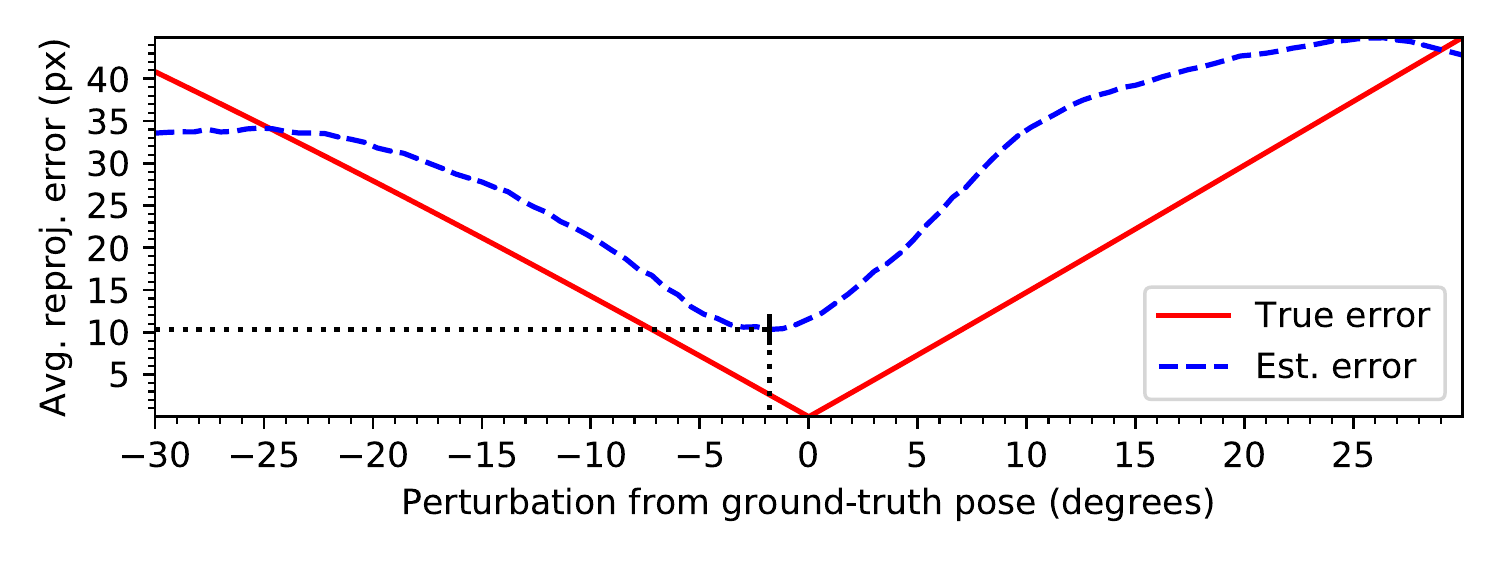}
    \vspace{-0.150\newheight}
    }

    \end{minipage}

    \vspace{-0.15cm}
    \caption{Estimated average reprojection error of our network for the \emph{cat} object in two test frames of \lmo{} \cite{Brachmann_2014_ECCV}. A rotational perturbation is applied for a fixed axis in the camera coordinate frame, in the range of $[-30,30]$ degrees. The estimated minimum is marked in the figure. (a-b) An unoccluded example. (c-d) An example with occlusion.}
    \label{fig:est_error_examples}
\end{figure}

Shared among contemporary rendering-based pose refinement methods \cite{Manhardt_2018_ECCV,li2019ijcv,Zakharov_2019_ICCV} is the approach of feeding an observed as well as a synthetically rendered image as input to a CNN model, which is trained to predict the relative pose of an object between the two images.
Our key insight is that rendering-based pose refinement is possible without explicitly regressing to the parameter vector of the relative pose.
Instead, estimating an error function of the relative pose is enough, since minimization of said error function w.r.t.\ pose can be done during inference.
Figure \ref{fig:est_error_examples} shows error function estimates for two test frames.
Although a larger bias is observed in the occluded case, the estimated minimum is still close to ground-truth.
Our main contributions can be summarized as follows:
\begin{itemize}
    \item A novel pose refinement method, which works well without real training data.
    \item Robustness to partial occlusions, by the implicit nature of our method, making it insensitive to over- or under-estimations of the error function.
    \item State-of-the-art results for two out of three metrics on \lmo{} \cite{Brachmann_2014_ECCV}.
\end{itemize}

Our pose refinement pipeline takes as input (a) an object CAD model and (b) an initial pose estimate, referred to as a ``pose proposal''.
The pose proposal is assumed to be obtained from another method, and is fed to the refinement pipeline, consisting of three parts:
\begin{enumerate}
    \item Synthetic rendering of the detected object under the pose proposal.
    \item Estimation of the average reprojection error of all model points, when projected into the image using the ground truth pose as well as the pose proposal.
    \item Iterative refinement of the 6D pose estimate by minimizing the reprojection error. %
\end{enumerate}
We will refer to our method as Pose Proposal Critic (PPC), since the heart of the method involves judging the quality of a pose proposal.

\section{Related Work}

Similarly to the work of Kendall \etal{} for camera localization \cite{Kendall_2015_ICCV}, the rigid object pose estimation methods of Xiang \etal{} \cite{xiang2017posecnn} and Do \etal{} \cite{do_deep6d} estimate the pose by directly regressing the pose parameters.
The most successful methods for rigid object pose estimation do however make use of a two-stage pipeline, where 2D-3D correspondences are first established, and the object pose is then retrieved by solving the corresponding camera resectioning problem \cite{hartley-zisserman-book-2004}.
A common approach has been to regress 2D locations of a discrete set of object keypoints, projected into the image, yielding a sparse set of correspondences \cite{Rad_2017_ICCV,Kehl_2017_ICCV,Tekin_2018_CVPR}.
Other methods instead output heatmaps in order to encode said keypoint locations \cite{pavlakos_icra_2017,Oberweger_2018_ECCV}.

Among the sparse correspondence methods, Oberweger \etal{} \cite{Oberweger_2018_ECCV} stand out in that they address the problem of partial occlusions very carefully. They show that occluders typically have a corrupting effect on CNN activations, far beyond the occluded region itself, and that training with occluded samples might not help to overcome this problem. Instead they resort to limiting the receptive field of their keypoint detector, as a crude but effective way to limit the impact of occluders.

Dense correspondence methods on the other hand, are inherently more robust to large variances in correspondence estimates.
Pixel-wise regression on the corresponding (object frame) 3D coordinates has been proposed by \cite{ipose_accv18,Park_2019_ICCV,Li_2019_ICCV}, while Zakharov \etal{} \cite{Zakharov_2019_ICCV} take a different approach and discretize the object surface into smaller segments, and then perform classification on which segment is visible in which pixel.
Hu \etal{} \cite{Hu_2019_CVPR} use a sparse set of keypoints, but yet leverage on dense correspondences due to redundantly regressing a number of 2D locations for each object keypoint.
Peng \etal{} \cite{Peng_2019_CVPR} take a similar approach, but simplify the output space by regressing to the direction from each pixel to each of the projected keypoints, but not the corresponding distance. Pair-wise sampling of pixel-wise predictions then yields votes for keypoint locations.

The method of Peng \etal{} \cite{Peng_2019_CVPR} does indeed prove  robust to partial occlusions, and yields accurate estimates of rotation as well as lateral translation on the \lmo{} benchmark.
Consequently, the results are state-of-the-art for the depth-insensitive metric based on reprojection errors.
Nevertheless, their depth estimates are still not accurate, and suffer in the presence of partial occlusion.
The rendering-based pose refinement method of DeepIM (Li \etal{} \cite{li2019ijcv}) does however perform well on \lmo{} for all common metrics, and in particular gives a huge boost in depth estimation accuracy, yielding state-of-the-art results for the metric based on matching point clouds in 3D, and suggesting that rendering-based pose refinement is a powerful tool for accurate pose estimation in the presence of partial occlusion. We will experimentally compare to DeepIM and show how one can achieve significantly improved results for partial occlusions.

Moreover, we point out that while a multitude of approaches for increasing robustness, especially to partial occlusions, has been observed among correspondence-based pose estimation methods, we have not yet seen any directed efforts to address these issues in the literature of rendering-based pose refinement.

When it comes to rendering-based pose refinement, early work was done by Tjaden and Schömer \cite{Tjaden_2017_ICCV}, proposing a segmentation pipeline based on hand-crafted features, and iterative alignment of silhouettes. Rad and Lepetit \cite{Rad_2017_ICCV} also apply rendering-based refinement as part of the BB8 pose estimation pipeline, improving on initial estimates. BB8 is based on sparse correspondences ($8$ bounding box corners),
and a refinement CNN is trained to regress the reprojection errors for each of the bounding box corners. Refinement is then carried out on the correspondences themselves, yielding an updated camera resectioning problem to be solved. In contrast, our method instead estimates the average reprojection error over all model points and refinement is done directly on the pose. %

Manhardt \etal{} \cite{Manhardt_2018_ECCV}, Li \etal{} \cite{li2019ijcv} and Zakharov \etal{} \cite{Zakharov_2019_ICCV}, all propose a CNN-based refinement pipeline, where the model is trained to learn the relative pose between an observed image and a synthetically rendered image under a pose proposal.
The main difference between their approaches and ours is that we instead choose to learn an error function of the relative pose.
Among these methods, \cite{li2019ijcv} is the only one that handles partial occlusions well.
The results of \cite{Zakharov_2019_ICCV} seem competitive at a first glance, but evaluation is only carried out on the frames for which the 2D object detector successfully detected the object of interest, and furthermore parts of the \lmo{} dataset were used for training, which does not allow for a fair comparison.

\section{Method}

In this section, the three main parts of our %
pipeline will be described in detail.

The core idea of our approach is that even though neural networks have an amazing capacity to learn difficult estimation tasks, the learning problem should be kept as simple as possible. Given a pose proposal, the task of our network is to determine how good the proposal is with respect to the ground truth. So, instead of trying to learn the pose parameters directly, it is only required for the network to act as a critic of different proposals. To further simplify the task, we render a synthetic image using the pose proposal, and then the network only needs to determine if the \emph{rendered} image is similar to the \emph{observed} image or not. As a measure of similarity, we use the average reprojection error of object CAD model points. Then, at inference, the objective is to find the pose parameters with lowest predicted reprojection error, resulting in a minimization problem which can be solved with standard optimization techniques.

It is assumed that intrinsic camera parameters are known for the \emph{observed} images, and that a three-dimensional CAD model of the object of interest is available.

\subsection{Part I: Rendering the Object Under a Pose Proposal}
Similar to previous work \cite{Rad_2017_ICCV,li2019ijcv,Manhardt_2018_ECCV,Zakharov_2019_ICCV}, we render a synthetic image of a detected object based on the suggested pose proposal.
Rendering is done
on the GPU using OpenGL with Lambertian shading and the light source at the camera center. The background is %
kept black.

Rather than using all of the \emph{observed} image directly, we zoom in on the detected object. Zooming is done based on the current pose proposal yielding square image patches centered at the projection of the object center. The size of the corresponding image patches (\emph{observed} and \emph{rendered}) is chosen as $1.2$ times the projection of the object diameter, and furthermore, the \emph{observed} patch is bilinearly upsampled to $512\times 512$ pixels.
Note that the object will be centered in the \emph{rendered} image patch, but need not be centered in the \emph{observed} image unless the pose proposal is accurate.

For future reference, let $\mathcal{Z}_\theta$ denote the zoom-in operator for pose proposal $\theta$, acting on \emph{observed} image, $I_{obs}$, resulting in image patch $P_{obs}=\mathcal{Z}_\theta I_{obs}$. We will denote the \emph{rendered} image patch by $P_{rend}$. For performance reasons, the patch is rendered at $256\times 256$ resolution, and then bilinearly upsampled to $512\times 512$.

\subsection{Part II: Learning Average Reprojection Error}
We use a pretrained optical flow network as backbone, add a regression head and finetune in order to take the \emph{observed} and \emph{rendered} image patches as input, and output an estimate of the average reprojection error, i.e., the average image distance between the projected CAD model points using the ground truth and the pose proposal, respectively.

Let $f = f(\mathcal{Z}_\theta I_{obs}, P_{rend}(\theta))$ be the estimated error of
the neural network, where $\mathcal{Z}_\theta I_{obs}$ is the \emph{observed} image patch and $P_{rend}(\theta)$ is the \emph{rendered} image patch for pose proposal $\theta$. If $\mathcal{P}_\theta$ denotes the projection of a 3D point onto the image patch using pose $\theta$\footnote{Note that the projection operator itself depends on the pose, due to the dependence of the zoomed-in image patch, and thus the effective intrinsic camera parameters, on the estimated object position.}, the reprojection error to be estimated by the network $f$ is then given by
\begin{equation}
    \frac{1}{M} \sum_{i=1}^M \left|\left| \mathcal{P}_{\hat \theta}(R_{\hat \theta} p_i + t_{\hat \theta}) - \mathcal{P}_{\hat \theta}(R_{\theta^*} p_i + t_{\theta^*}) \right|\right|_2 ,
    \label{eq:reproj}
\end{equation}
where $\hat \theta$ and $\theta^*$ are the estimated and true poses, respectively, and $p_i$ are the $M$ object model points. Figure \ref{fig:est_error_examples} shows the estimated error function for two test frames of \lmo{}, one in which the object is partially occluded.

The reprojection error is measured in image patch pixels, i.e.\ after zoom-in rather than before.
Estimating the reprojection error before zoom-in would require the network to estimate and rescale with the absolute depth, which would introduce an unnecessary complication. The reason we choose the reprojection error is that we expect it to be relatively easy to infer from image pairs without a lot of high level reasoning, and thus providing a relatively easy learning task.
Furthermore, the reprojection error is quite related to optical flow, and should fit particularly well with a pretrained optical flow backbone.

For further details on the implementation, we refer to the appendix.
Section~\ref*{sec:network_arch} gives more details on the network architecture and Section~\ref*{sec:training_and_loss_fcn} on the loss function and hyperparameters used for training.
How to sample the pose proposals during training is covered in Section~\ref*{sec:proposal_sampling}, while Section~\ref*{sec:synth_data_aug} describes data augmentation strategies, and in particular how to generate synthetic training examples.

\subsection{Part III: Minimizing Reprojection Error}
Once the CNN $f$ is trained for estimating reprojection errors, we may apply it for refining an initial pose proposal $\theta_0$ of our object of interest in the \emph{observed} image $I_{obs}$.

Let the compound function $J(\theta) = f(\mathcal{Z}_\theta I_{obs}, P_{rend}(\theta))$ encapsulate the operations of rendering, zoom-in and the CNN itself, which leads to the optimization problem: $\min_\theta J(\theta)$.
We minimize $J$ locally, initializing at $\theta_0$.
Gradient-based optimization is carried out and although analytical differentiation is a tempting approach in the light of differentiable renderers such as \cite{Kato_2018_CVPR}, we observed noisy behavior in $J$, and we instead apply numerical differentiation for %
robustly estimating $\nabla_\theta J$.

For parameterizing the rotation, we take advantage of the Lie Algebra of $\SO{3}$. The initial rotation $R_0$ is used as a reference point and the parameterization is $R(\theta_r) = e^{A(\theta_r)} R_0$, where the parameters $\theta_r$ constitute the three elements of the $3\times 3$ skew-symmetric matrix $A(\theta_r)$. %
The translation is split into two parts.
The lateral translation $\theta_l$ represents the deviation from the projection of the initial position in pixels, i.e.\ $\mathcal{P}_{\theta_0}(t)-\mathcal{P}_{\theta_0}(t_0)$, where $\theta_0$ denotes the initial pose proposal, and $t_0$ denotes the translation part specifically.
The depth is parameterized as $d(\theta_d)=e^{\theta_d} d_0$, where $d_0$ is the initial depth estimate.

\subsubsection{Optimization Scheme}
It needs to be stressed that there may be spurious local minima in $J$, and for this reason care should be taken during the optimization. For better control over the procedure, we let decoupled optimizers run in parallel for the rotation / depth / lateral translation parameters, with different hyperparameters and step size decay schedules.
We apply in total $100$ iterations.

In order to handle non-convex and noisy behavior of $J$, we use stochastic optimization. In particular, the Adam optimizer \cite{kingma_iclr_2015} proved effective for handling the fact that $J$ may be quite steep in the vicinity of the optimum, yet quite flat farther away from the optimum. This property did otherwise risk the optimizer taking too far steps when encountering "steep" points or easily getting stuck in local minima, if the step size was too high or low, respectively.

Consider for now the optimization w.r.t.\ rotation and depth.
The optimization is roughly carried out in two phases, first w.r.t.\ rotation and then depth, with a smooth transition between the two. This sequential strategy is due to two reasons: (1) Although optimization w.r.t.\ rotation works well despite a sub-optimal depth estimate, keeping $\theta_d$ fixed reduces noise for the moment estimates of the optimizer. (2) For precise depth estimation, a good estimate of the other parameters is crucial. The reprojection error to be estimated, as well as the \emph{rendered} image itself, is much less sensitive to depth perturbations than to the other pose parameters, and focusing specifically on $\theta_d$ in the final stage helped to improve depth estimation.

Optimization w.r.t.\ lateral translation proved relatively easy, and less coupled with the other parameters, i.e.\ a reasonable minimum may be found despite e.g.\ a poor rotation estimate. Particularly fast convergence of the lateral translation is desirable for the converse reason, that optimization w.r.t.\ the other parameters is coupled with the lateral translation estimate, and may not work well unless this is adequate. Luckily, convergence of these parameters is achieved in just a couple of iterations when using a plain SGD optimizer with momentum $0.5$ rather than Adam. The step size w.r.t.\ $\theta_l$ is set constantly to $1$.

\begin{figure}
  \centering
  \begin{minipage}{0.8\textwidth}

    \centering
    \includegraphics[width=0.9\textwidth,trim={0 0.4cm 0 0},clip=true]{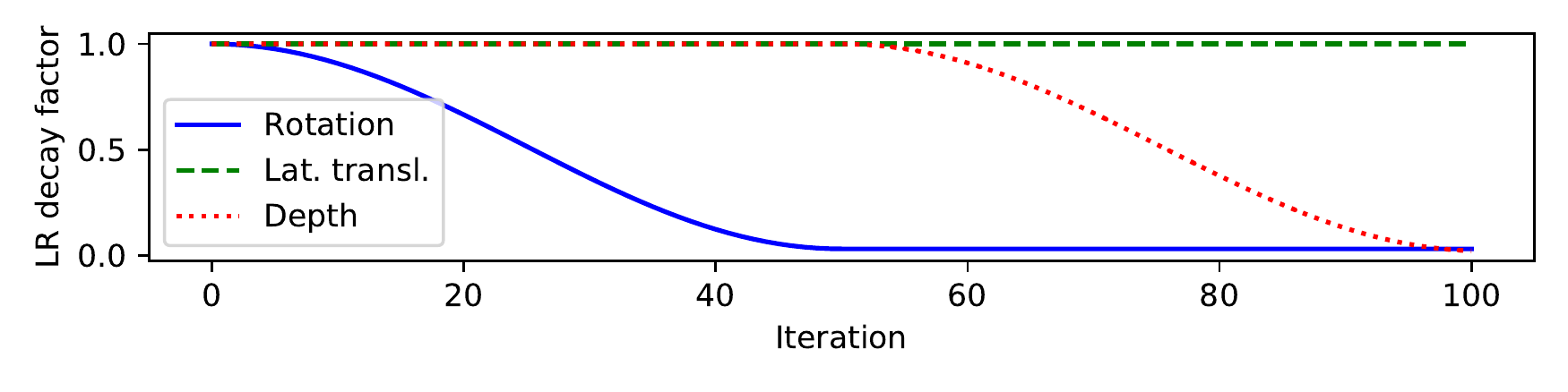}

  \end{minipage}
    \caption{Step size decay schedules for the different parameters $\theta_r$, $\theta_l$ and $\theta_d$ during inference. The decay value is relative to the respective initial step sizes.}
    \label{fig:lr_schedule}
\end{figure}

The step size decay schedule for all parameters is illustrated in Figure~\ref{fig:lr_schedule} and the exponential decay rates for the moment estimation of Adam were set to $(\beta_1, \beta_2) = (0.6, 0.9)$ for $\theta_r$, and $(\beta_1, \beta_2) = (0.4, 0.9)$ for $\theta_d$.
The step sizes used for finite differences were $0.01$, $1.0$ and $0.005$ for $\theta_r$, $\theta_l$, $\theta_d$, respectively.

\section{Experiments}
\subsection{Datasets and Training Data}
Experiments are carried out on \lm{} as well as \lmo{}.

\lm{} is a standard benchmark for rigid object pose estimation and was introduced by Hinterstoisser \etal{} \cite{Hinterstoisser2013}.
The dataset consists of $15$ object CAD models along with $15$ RGB-D image sequences of an indoor scene where objects are laid out on a table with cluttered background. For each sequence there is a corresponding object of interest put at the center.
Although depth images are provided, it is also a common benchmark for RGB-only %
pose estimation.
As two of the objects suffer from low quality %
CAD models, they are commonly excluded from evaluation and we follow the same practice.

The \lmo{} dataset was produced by Brachmann \etal{} \cite{Brachmann_2014_ECCV} by taking one of the \lm{} sequences and annotating the pose of the surrounding $8$ objects. While the central object is typically unoccluded, the surrounding objects are often partially occluded, resulting in a challenging dataset.
The %
central object is not part of the %
benchmark.

For experiments on \lmo{} we use real images from \lm{}, as has conventionally been done in the literature, while the \lmo{} images are only used for testing.
With $33~\%$ probability we sample a real training image, and with $67~\%$ probability a synthetic one. In the synthetic case, there is a $50~\%$ probability that $2$ occluding objects are rendered, and a $50~\%$ probability that no occluders are rendered. We carry out additional experiments on \lmo{} where the model is trained only on synthetic data, still with $50~\%$ of the samples being occluded.

For experiments on \lm{}, we split training and test data exactly as \cite{li2019ijcv}, with the same $\sim 200$ samples for training and $1000$ samples for test.
With $50~\%$ probability we sample a real training image and with $50~\%$ probability a synthetic one, but without any occluders.

The objects known as \texttt{eggbox} and \texttt{glue}, present in both datasets, are conventionally considered symmetric w.r.t.\ a $180$ degree rotation, but it can be argued whether these are considered actual symmetries.
Nevertheless, for these objects we duplicate the initial pose proposals with their $180$ degree rotated equivalents and %
refine the pose using both initializations. %
In the end, the iterate with the least estimated error is chosen.

\subsection{Evaluation Metrics for Pose Refinement}
For evaluation of our pose refinement method, we use three conventional metrics, explained in the following.
All of them are defined as the percentage of annotated object instances for which the pose is correctly estimated, i.e., the recall according to the specific ways of quantifying the error.

The average distance metric \addnosymm{} \cite{Hinterstoisser2013} is the percentage of object instances for which the object point cloud, when transformed with the estimated pose as well as the ground-truth pose, has an average distance less than $10~\%$ of the diameter of the object. The \addsymm{} metric \cite{Hinterstoisser2013} is closely related, and only differs in that the closest point distance is used, rather than the distance between corresponding points. In general \addnosymm{} is used, but \addsymm{} is used for objects that are considered symmetrical, and we let \add{} refer to the two of them together.
The \reproj{} metric is similar to \addnosymm{}, but differs in that the transformed point clouds are projected into the image before the mean distance is computed. The acceptance threshold is set to 5 pixels.
Finally, the \degcm{} metric accepts a pose estimate if the rotational and translational components differ from their ground-truth equivalents by at most $5$ degrees and $5$ cm, respectively.

\subsubsection{"Symmetric" Objects and Faulty Annotations}\label{sec:symm_objs_faulty_anno}
When it comes to the \reproj{} and \degcm{} metrics, they do typically not take symmetries into account. This is not a huge problem for the \lm{} dataset, partly because the high correlation between training and test data may help resolve any potential symmetries, and partly because, as pointed out earlier, none of the objects are truly symmetrical.

For \lmo{} however, the \texttt{eggbox} object is unfortunately annotated according to the supposedly equivalent $180$ degrees rotated pose, in all but the first $396$ frames. For this reason, above mentioned metrics make little sense.
Li \etal{} \cite{li2019ijcv} do however modify these metrics in order to evaluate against the most beneficial of all proposed symmetries, which makes much more sense given the circumstances. We follow their proposal and perform the evaluation on \lmo{} w.r.t.\ these symmetrically aware metrics, which we will refer to as \reprojs{} and \degcms{}.

\subsection{Pose Refinement Results}

Here we present our main pose refinement results.
For a comparison of different backbone networks and detailed per-object results, we refer the reader to Section~\ref*{sec:backbone_comparison}
and
Section~\ref*{sec:detailed_refinement_results}
in the appendix.
Illustrations of refinement iterates are also available, in Section~\ref*{sec:refinement_illustration}
.

State-of-the-art comparisons on the \lmo{} dataset are given in Table~\ref{tab:real_lmo_allmetrics}.
DeepIM~\cite{li2019ijcv} used initializations from PoseCNN~\cite{xiang2017posecnn}, but as these predictions are not publicly available,
we instead rely on initial pose proposals from PVNet~\cite{Peng_2019_CVPR}.
The evaluation of PVNet was carried out by us and is based on the \texttt{clean-pvnet} implementation along with pre-trained models\footnote{
At times we observed negative depth estimates from PVNet, which was corrected for according to Section~\ref*{sec:neg_label_correction} in the appendix.}.
Note that although the symmetry-aware \reprojs{} metric should be used on \lmo{} (see Section~\ref{sec:symm_objs_faulty_anno}), Oberweger \etal{} \cite{Oberweger_2018_ECCV} report their results based on the \reproj{} metric.
We also want to mention that CDPN~\cite{Li_2019_ICCV} perform well on \lmo{}, but no quantitative numbers are reported.
\begin{table}[h]
    \centering
    \begin{tabular}{|r|c|c|c|c|}
        \cline{2-5}
        \multicolumn{1}{c|}{} &
            \thead{Oberweger \etal{} \cite{Oberweger_2018_ECCV}} &
            \thead{PVNet \cite{Peng_2019_CVPR}} &
            \thead{PoseCNN \cite{xiang2017posecnn}\\ + DeepIM \cite{li2019ijcv}} &
            \thead{PVNet \cite{Peng_2019_CVPR}\\ + PPC (Ours)} \\
        \hline
        \add{} &
            \score{30,40} &
            \score{41,37} &
            \scorebf{55,50} &
            \score{55,33}\\
        \hline
        \reprojs{} &
            \score{60,86} &
            \score{61,84} &
            \score{56,61} &
            \scorebf{66,37}\\
        \hline
        \degcms{} &
            -- &
            \score{33,36} &
            \score{30,93} &
            \scorebf{41,52}\\
        \hline
    \end{tabular}
    \caption{Results on \lmo{}. Note that \cite{Oberweger_2018_ECCV} report results according to the \reproj{} metric instead of \reprojs{}.}
    \label{tab:real_lmo_allmetrics}
\end{table}

We also present results on the \lmo{} dataset where we train purely on synthetic data, see Table~\ref{tab:synth_lmo_allmetrics}. Our initial pose proposals are obtained from CDPN~\cite{Li_2019_ICCV}, which was
the previous state-of-the-art for this set-up (cf.\ 
Benchmark for 6D Object Pose Estimation (BOP) evaluation server \cite{hodan2018bop}).
Also note that for these experiments only a subset of $200$ test frames is used, in compliance with BOP.
\begin{table}[h]
    \centering
    \begin{tabular}{|r|c|c|c|c|}
        \cline{2-3}
        \multicolumn{1}{c|}{} &
            \thead{CDPN-synth \cite{Li_2019_ICCV}} &
            \thead{CDPN-synth \cite{Li_2019_ICCV}\\ + PPC-synth (Ours)} \\
        \hline
        \add{} &
            \score{18,76} &
            \scorebf{23,59} \\
        \hline
        \reprojs{} &
            \score{32,22} &
            \scorebf{35,99} \\
        \hline
        \degcms{} &
            \score{16,13} &
            \scorebf{19,81} \\
        \hline
    \end{tabular}
    \caption{Results on \lmo{} using only synthetic training data.}
    \label{tab:synth_lmo_allmetrics}
\end{table}

Finally, results on the \lm{} dataset are presented in Table~\ref{tab:real_lm_allmetrics} with the
purpose of providing a direct comparison with DeepIM~\cite{li2019ijcv} with identical initializations.
We outperform DeepIM on all metrics using the same proposals from PoseCNN~\cite{xiang2017posecnn}.
Note that although no results on \lm{} for PoseCNN are reported in \cite{xiang2017posecnn}, predictions by PoseCNN are made available by~\cite{li2019ijcv}.
The results are also good when compared to the state-of-the-art pose estimation methods of Li \etal{} \cite{Li_2019_ICCV} and Peng \etal{} \cite{Peng_2019_CVPR} on \lm{}.
\begin{table}[h]
    \centering
    \begin{tabular}{|r|c|c|c|}
        \cline{2-4}
        \multicolumn{1}{c|}{} &
            \thead{PoseCNN \cite{xiang2017posecnn}} &
            \thead{PoseCNN \cite{xiang2017posecnn}\\ + DeepIM \cite{li2019ijcv}} &
            \thead{PoseCNN  \cite{xiang2017posecnn}\\ + PPC (Ours)} \\
        \hline
        \add{} &
            \score{62,04} &
            \score{88,33} &
            \scorebf{88,67}\\
        \hline
        \reproj{} &
            \score{64,52} &
            \score{97,53} &
            \scorebf{97,60}\\
        \hline
        \degcm{} &
            \score{18,14} &
            \score{85,21} &
            \scorebf{89,74}\\
        \hline
    \end{tabular}
    \caption{Comparison with the refinement method of DeepIM and ours on \lm{} with PoseCNN as initialization. Note that \cite{li2019ijcv} reports results according to \reprojs{} and \degcms{} metric instead of \reproj{} and \degcm{}.}
    \label{tab:real_lm_allmetrics}
\end{table}

\subsection{Running Time}
Experiments were run on a workstation with 64 GB RAM, Intel Core i7-8700K CPU, and Nvidia GTX 1080 Ti GPU.
Our pose refinement pipeline takes on average 33 seconds per frame during inference for the $100$ iterations to be carried out, meaning 3 iterations / s.
One way to improve on this could be by enabling analytical differentiation through differentiable rendering, although care should be taken in order to make sure that $J(\theta)$ behaves smoothly enough, for instance, with a regularization scheme.
Furthermore, rather than using iterative gradient-based optimization, gradient-free and sample-efficient approaches such as Bayesian optimization could be worth exploring, but is left as future research.

\section{Conclusion}
We have presented a novel rendering-based pose refinement method, which shows improved performance compared to previous refinement methods, and is robust to partial occlusions.

On the \lmo{} benchmark, we initialize our method with pose proposals from PVNet \cite{Peng_2019_CVPR}, yielding state-of-the-art results for two out of three metrics on this competitive benchmark, while performing on-par with previous methods for the third metric.
Furthermore, 
 additional experiments on \lmo{} show that our method works well also when trained purely on synthetic data, improving on the pose estimates of CDPN~\cite{Li_2019_ICCV}.
Finally, on the \lm{} benchmark, previous refinement methods are outperformed for all metrics.

\section{Acknowledgements}
This work was partially supported by the Wallenberg AI, Autonomous Systems and Software Program (WASP) funded by the Knut and Alice Wallenberg Foundation.

\clearpage
\appendix
\section{Implementation Details}
\subsection{Network Architecture}\label{sec:network_arch}
Similar to \cite{li2019ijcv}, we use a pretrained FlowNetSimple optical flow network as backbone.
They used the original model from Dosovitskiy \etal{} \cite{Dosovitskiy_2015_ICCV}, while we use the FlowNet 2.0 version from Ilg \etal{} \cite{Ilg_2017_CVPR}.
We flatten the encoder output feature maps, and feed them through three fully-connected layers, constituting our main branch. In contrast to \cite{li2019ijcv}, we use standard ReLU rather than leaky-ReLU activation functions. Furthermore, for the hidden layers we use $1024$ neurons, and apply dropout with $30~\%$ probability.
The final layer outputs one neuron, representing the average reprojection error estimate.

We also follow \cite{li2019ijcv} in the approach of adding an auxiliary branch for foreground / background segmentation, by adding a 1-channel $3\times 3$ convolutional layer next to the optical flow prediction at level 4 (a.k.a.\ \texttt{flow4}). The optical flow  prediction itself is however disregarded in order to simplify the pipeline, and we point out that the ablation study of \cite{li2019ijcv} showed only a minor boost from including this auxiliary task.

Finally, the FlowNet 2.0 network is fed only the \emph{observed} and \emph{rendered} image patches as inputs, and no segmentation is provided, as for \cite{li2019ijcv}. Experiments on re-training their network without segmentation input did however not result in any performance drop.

\subsection{Training and Loss Function}\label{sec:training_and_loss_fcn}
We trained on the loss function $\mathcal{L} =  0.01\cdot \mathcal{L}_{reproj} + 0.3\cdot \mathcal{L}_{seg}$, where $\mathcal{L}_{reproj}$ is the $L_1$ error between the true and estimated average reprojection error, and  $\mathcal{L}_{seg}$ is the binary cross-entropy loss of the foreground / background segmentation, averaged over all pixels.
Furthermore, the target for average reprojection error was saturated at $50$ px, making sure that very large perturbation samples will not introduce a disturbance into the training, letting the network focus on achieving high precision within the range of reasonable perturbations.

One model was trained for each object, for $75$ epochs with $42$ batches sampled in each epoch. The learning rate was initialized to $5\cdot 10^{-5}$, and multiplied by $0.3$ every $10$ epochs. The batch size was $12$ and $L_2$ regularization was applied with weight decay $5\cdot 10^{-4}$.

\subsection{Pose Proposal Sampling}\label{sec:proposal_sampling}
For training, we generate pose proposals by perturbing the ground-truth pose in three different ways: 
(1) With $30~\%$ probability, a rotation around a random axis going through the object centre, whose magnitude is normally distributed with $\mu=0$ and $\sigma=45$ degrees.
(2) With $30~\%$ probability, a random lateral translation, normally distributed with $\mu=0$ and $\sigma=0.1d$, where $d$ is the object diameter.
(3) With $40~\%$ probability, a relative depth perturbation, sampled from a log-normal distribution with $\mu=0$ and $\sigma=\log 0.05$.
This procedure and settings were found to work experimentally well.

\begin{figure}
    \centering

    \subcaptionbox{}[0.23\textwidth]{
        \includegraphics[width=0.23\textwidth]{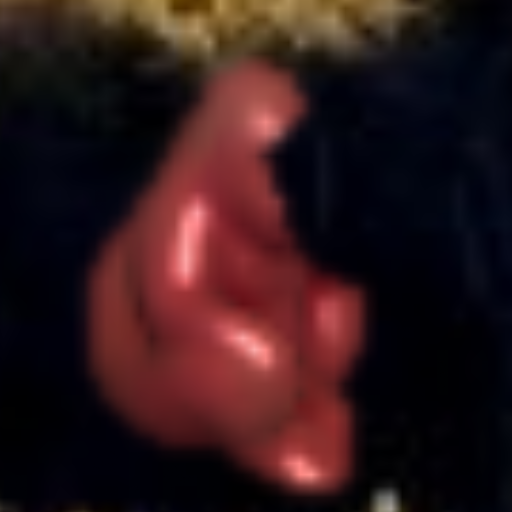}
    }
    \hfill
    \subcaptionbox{}[0.23\textwidth]{
        \includegraphics[width=0.23\textwidth]{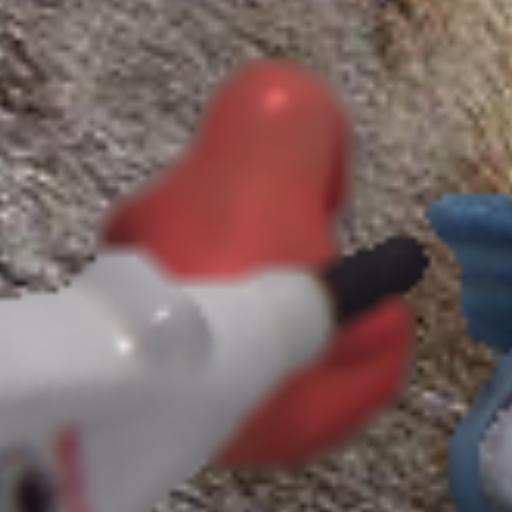}
    }
    \hfill
    \subcaptionbox{}[0.23\textwidth]{
        \includegraphics[width=0.23\textwidth]{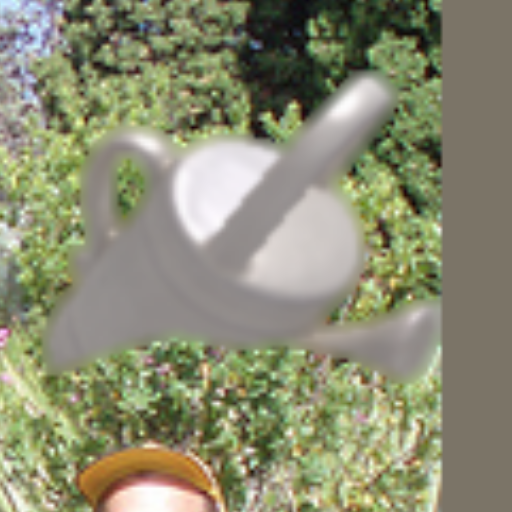}
    }
    \hfill
    \subcaptionbox{}[0.23\textwidth]{
        \includegraphics[width=0.23\textwidth]{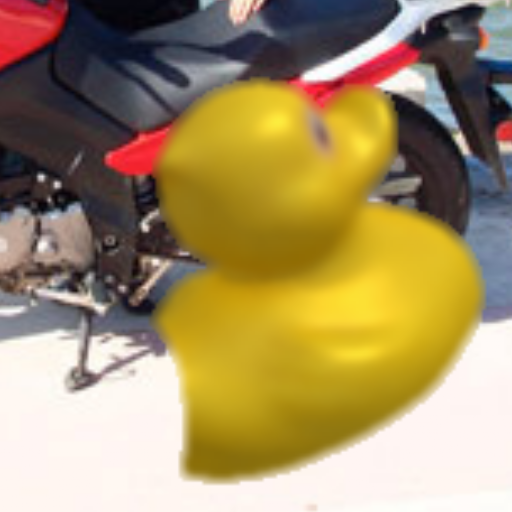}
    }\\

    \subcaptionbox{}[0.23\textwidth]{
        \includegraphics[width=0.23\textwidth]{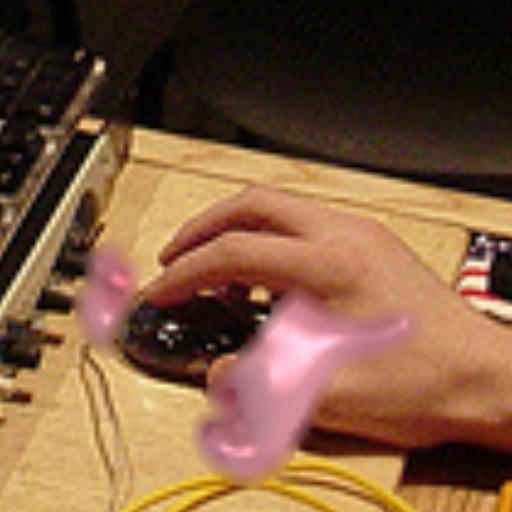}
    }
    \hfill
    \subcaptionbox{}[0.23\textwidth]{
        \includegraphics[width=0.23\textwidth]{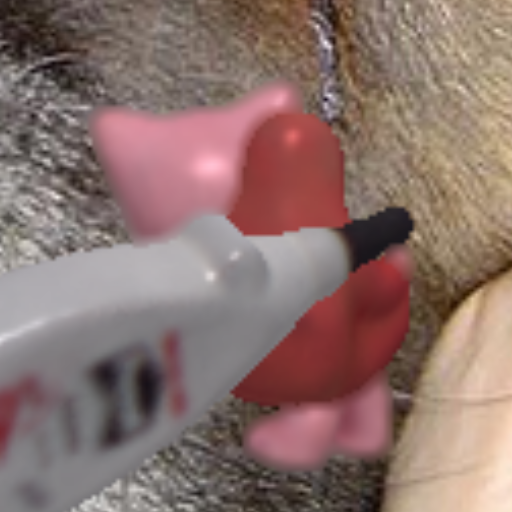}
    }
    \hfill
    \subcaptionbox{}[0.23\textwidth]{
        \includegraphics[width=0.23\textwidth]{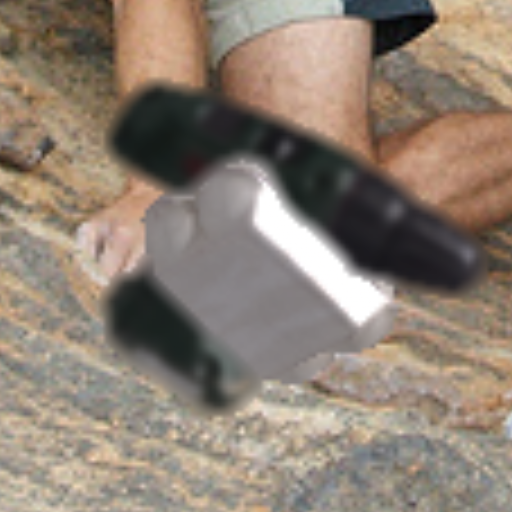}
    }
    \hfill
    \subcaptionbox{}[0.23\textwidth]{
        \includegraphics[width=0.23\textwidth]{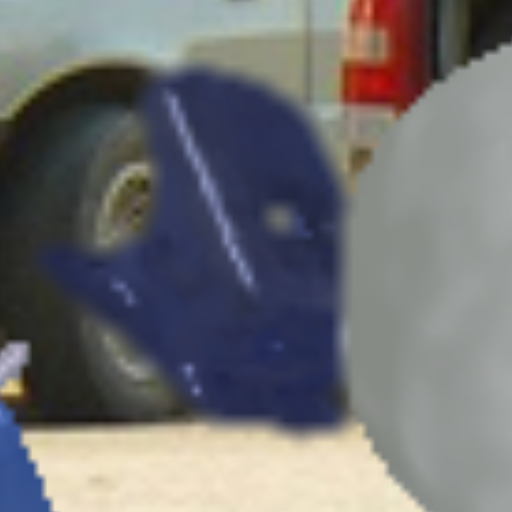}
    }

    \caption{Synthetically rendered training examples of \emph{observed} images. Occlusion is simulated by rendering additional objects in front of the object of interest, or alternatively the corresponding region is replaced with background, effectively making it transparent.}
    \label{fig:synth_train_imgs}
\end{figure}
\subsection{Rendering Synthetic Training Data}\label{sec:synth_data_aug}
In addition to real annotated training images, we augment the training examples by rendering synthetic \emph{observed} images, illustrated in Figure \ref{fig:synth_train_imgs}. Like for the pose proposals, rendering is done %
using OpenGL. Phong shading is applied and we noticed a performance boost by taking specular effects into account.
In the spirit of Domain Randomization \cite{tobin_2017_iros}, we sample variations in light source position as well as shading parameters such as ambient / diffuse / specular weights, and the whiteness / shininess parameters of the specular effects. No perturbations are applied on albedo.

Random images from Pascal VOC2012 \cite{pascal-voc-2012} were used as background and Gaussian blur was applied on the border in order to blend foreground and background and reduce overfitting to border artifacts as proposed by \cite{Dwibedi_2017_ICCV}. Gaussian blur was also applied to the whole object of interest as advised by \cite{Hinterstoisser_2018_ECCV_Workshops}.

Furthermore, occluding objects of other object categories are sometimes rendered in front of the object of interest. A visible region of at least $200$ pixels is however ensured, otherwise occluders are resampled. In order to prevent overfitting towards the specific objects used for occlusion, occluded regions are replaced with background with a $50~\%$ probability.

Finally, in the cases when we trained only on synthetic data, random noise in HSV-space was applied to the \emph{observed} images.

\section{Further Results}\label{sec:further_results}
\subsection{Backbone Comparison}\label{sec:backbone_comparison}
As a first experiment, we evaluated the performance when altering the backbone on \lmo{}.
In addition to using the FlowNet~2.0 backbone \cite{Ilg_2017_CVPR}, the encoder of Zakharov \etal{} \cite{Zakharov_2019_ICCV}, based on ResNet-18 \cite{He_2016_CVPR} and a siamese network was re-implemented for comparison.
As can be seen in Table~\ref{tab:real_lmo_backbone}, the FlowNet~2.0 model outperforms the alternative, giving further evidence for the conclusion made by Li \etal{} \cite{li2019ijcv} that a feature extractor trained for optical flow is useful also for this task.
\begin{table}
    \centering
    \begin{tabular}{|r|c|c|}
        \cline{2-3}
        \multicolumn{1}{c|}{} &
            \thead{ResNet-18 \cite{He_2016_CVPR}} &
            \thead{FlowNet 2.0 \cite{Ilg_2017_CVPR}} \\
        \hline
        \add{} &
            \score{50,92} & \scorebf{55,33} \\
        \hline
        \reprojs{} &
            \score{62,66} & \scorebf{66,37} \\
        \hline
        \degcms{} &
            \score{39,72} & \scorebf{41,52} \\
        \hline
    \end{tabular}
    \caption{Comparison of our method on \lmo{} for different backbones.}
    \label{tab:real_lmo_backbone}
\end{table}

\subsection{Illustration of Refinement Iterates}\label{sec:refinement_illustration}
\begin{figure}
    \centering
    \begin{minipage}{1.0\textwidth}

        \begin{minipage}{0.19\textwidth}
            \centering
            \includegraphics[width=\textwidth]{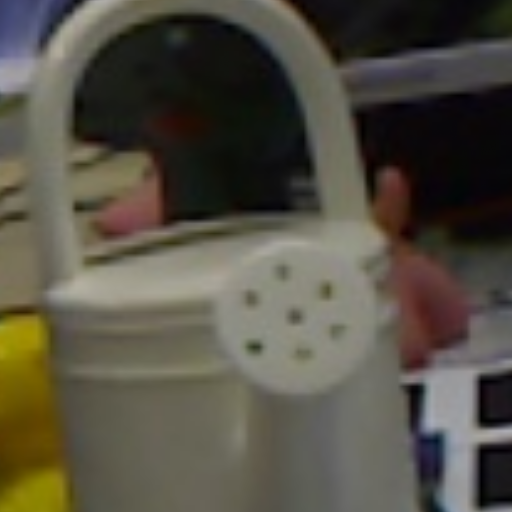}\\[0.05\textwidth]
            \includegraphics[width=\textwidth]{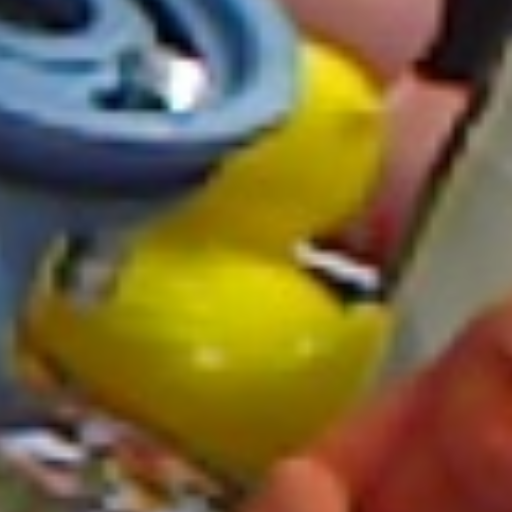}\\[0.05\textwidth]
            \includegraphics[width=\textwidth]{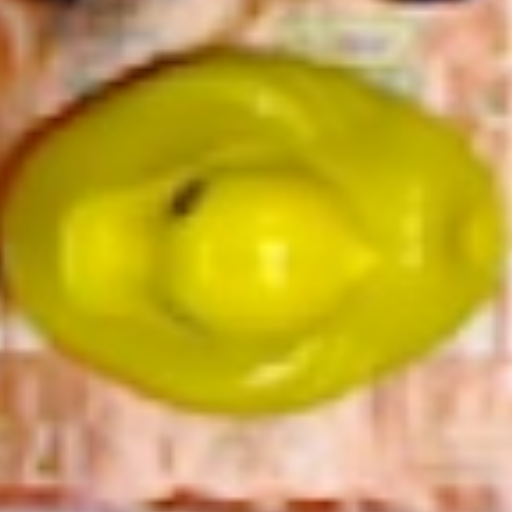}\\[0.05\textwidth]
            \includegraphics[width=\textwidth]{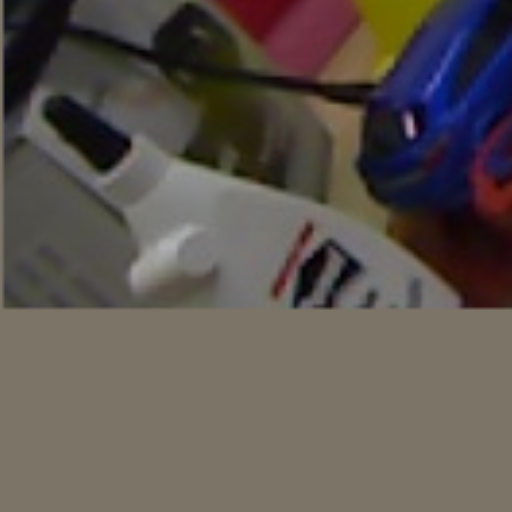}\\
            Initial \emph{obs}
        \end{minipage}
        \hfill
        \begin{minipage}{0.19\textwidth}
            \centering
            \includegraphics[width=\textwidth]{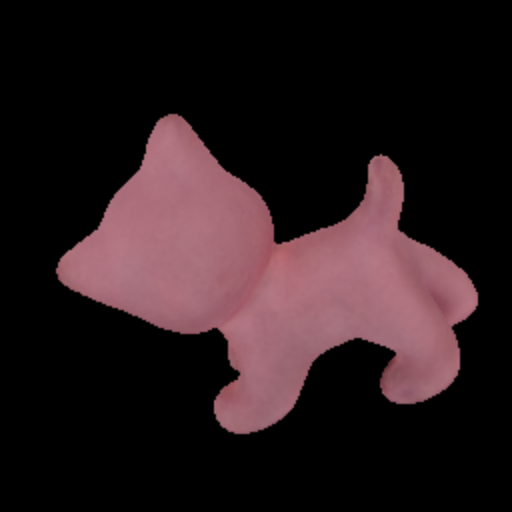}\\[0.05\textwidth]
            \includegraphics[width=\textwidth]{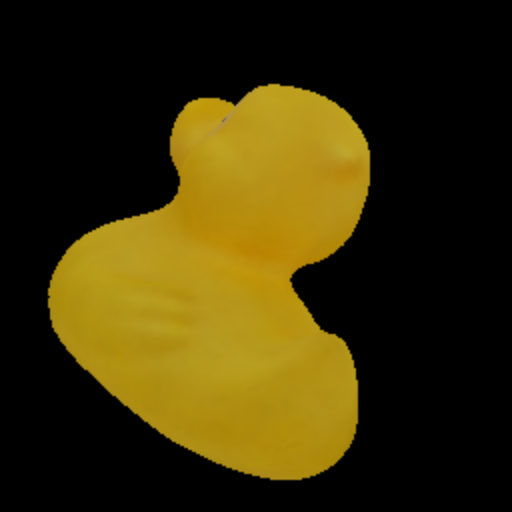}\\[0.05\textwidth]
            \includegraphics[width=\textwidth]{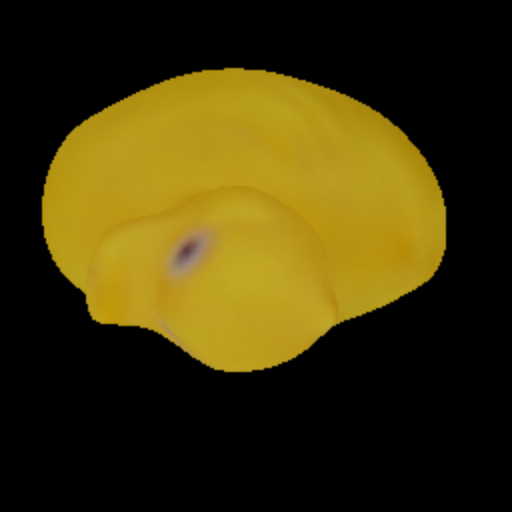}\\[0.05\textwidth]
            \includegraphics[width=\textwidth]{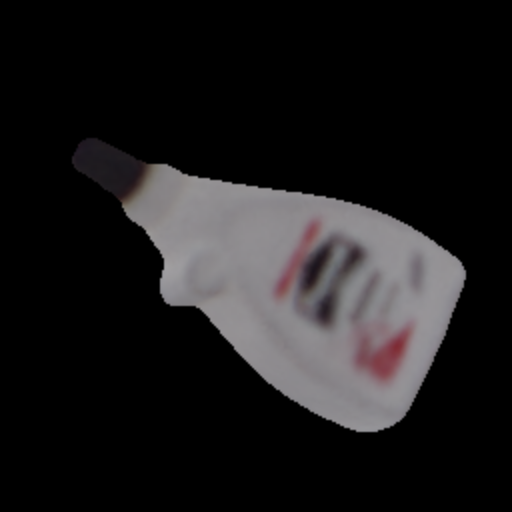}\\
            Initial \emph{rend}
        \end{minipage}
        \hfill
        \begin{minipage}{0.19\textwidth}
            \centering
            \includegraphics[width=\textwidth]{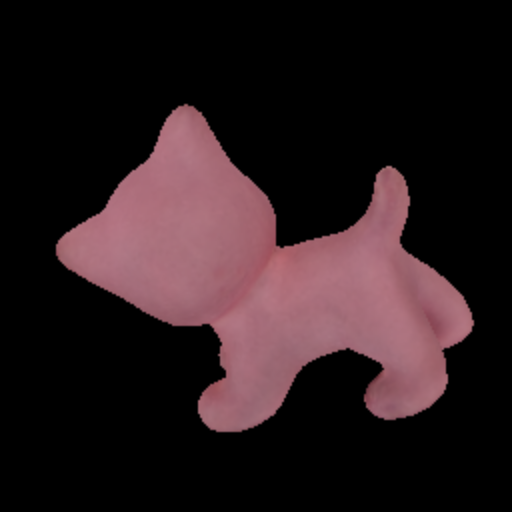}\\[0.05\textwidth]
            \includegraphics[width=\textwidth]{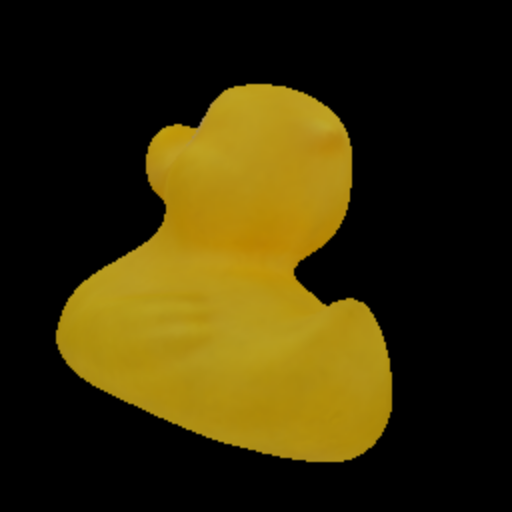}\\[0.05\textwidth]
            \includegraphics[width=\textwidth]{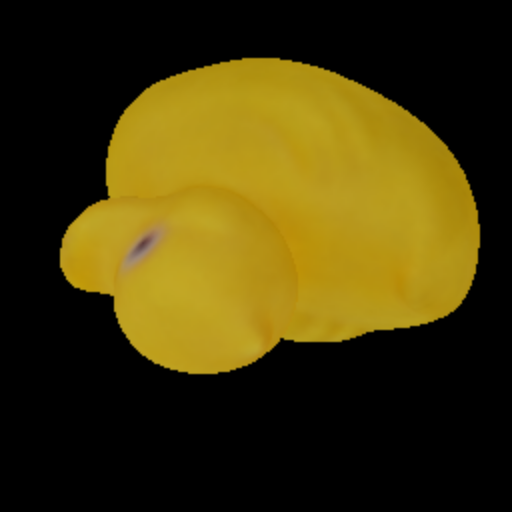}\\[0.05\textwidth]
            \includegraphics[width=\textwidth]{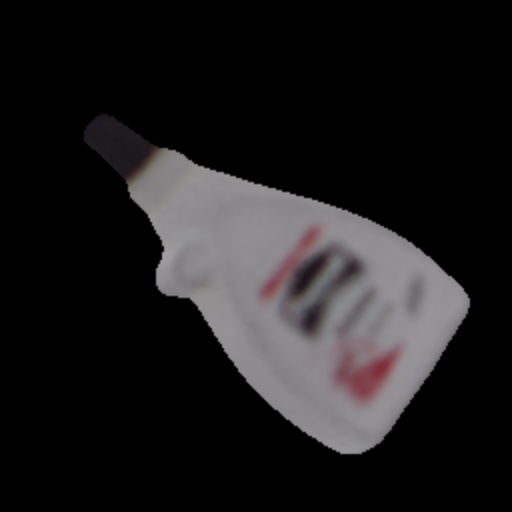}\\
            Iteration 10 \emph{rend}
        \end{minipage}
        \hfill
        \begin{minipage}{0.19\textwidth}
            \centering
            \includegraphics[width=\textwidth]{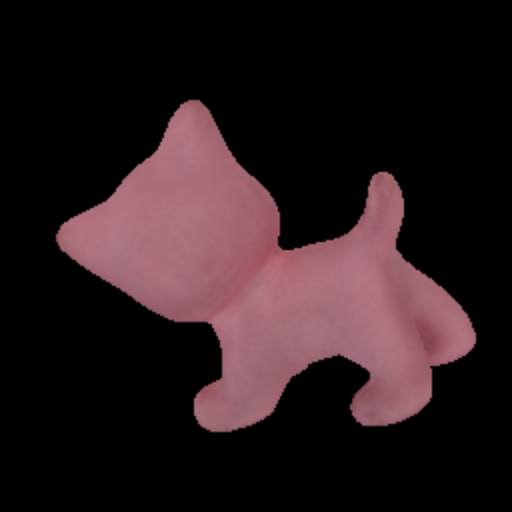}\\[0.05\textwidth]
            \includegraphics[width=\textwidth]{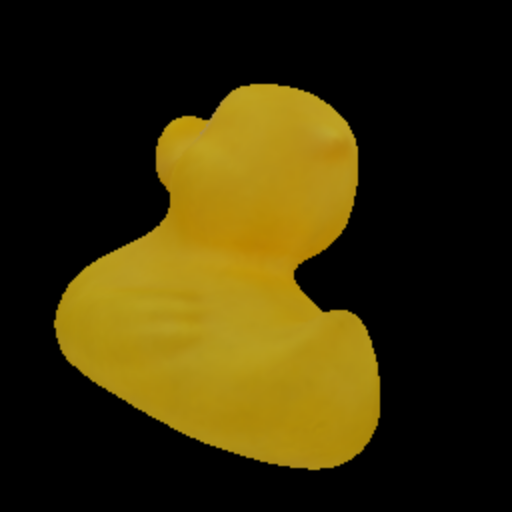}\\[0.05\textwidth]
            \includegraphics[width=\textwidth]{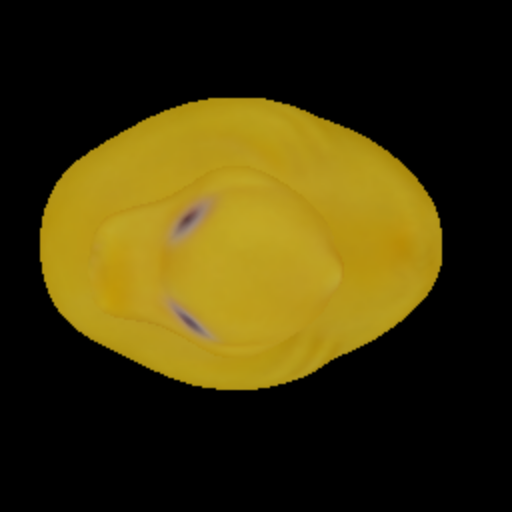}\\[0.05\textwidth]
            \includegraphics[width=\textwidth]{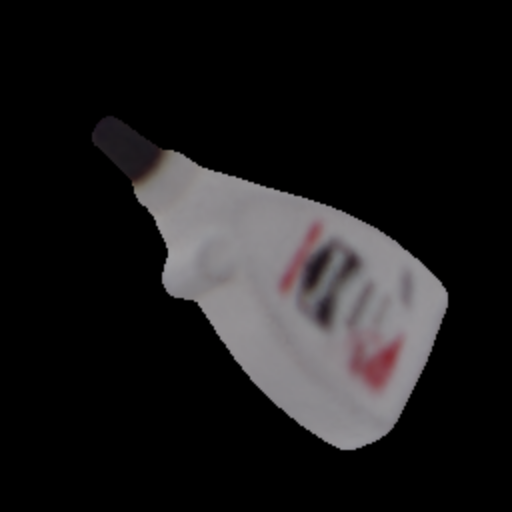}\\
            Final \emph{rend}
        \end{minipage}
        \hfill
        \begin{minipage}{0.19\textwidth}
            \centering
            \includegraphics[width=\textwidth]{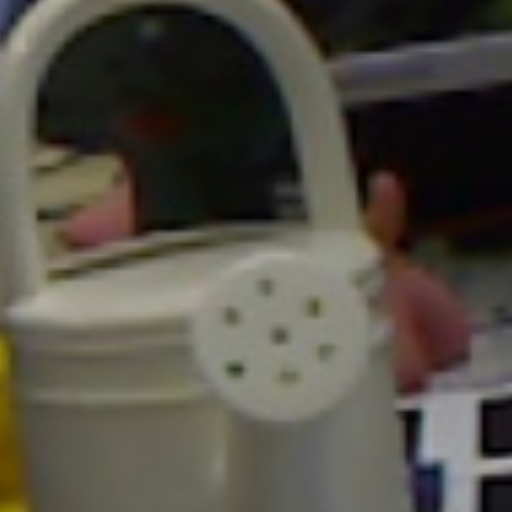}\\[0.05\textwidth]
            \includegraphics[width=\textwidth]{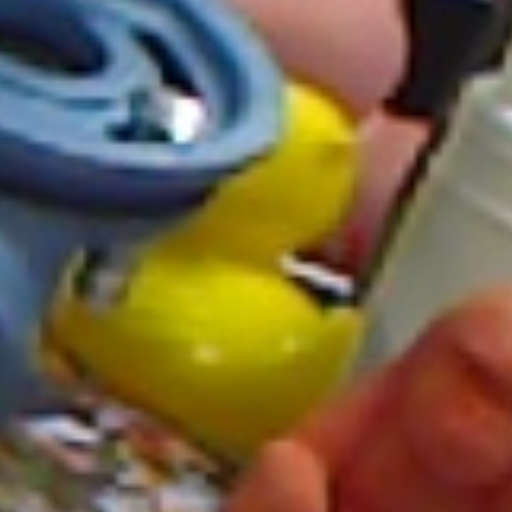}\\[0.05\textwidth]
            \includegraphics[width=\textwidth]{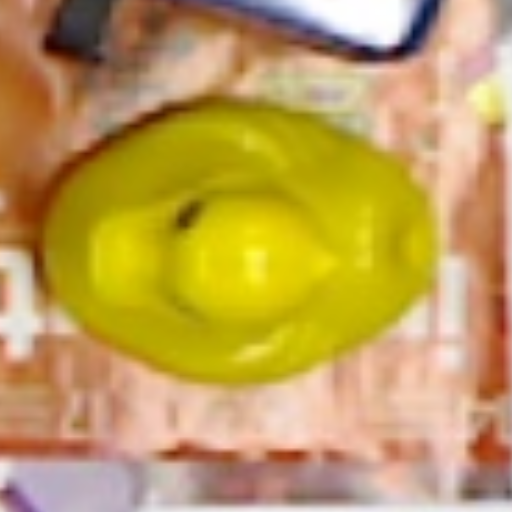}\\[0.05\textwidth]
            \includegraphics[width=\textwidth]{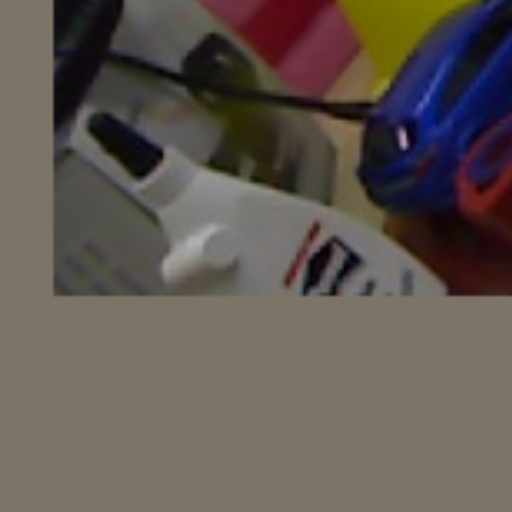}\\
            Final \emph{obs}
        \end{minipage}

    \end{minipage}

    \caption{Image patches during pose refinement iterations, for a few example frames of the \lmo{} dataset.}
    \label{fig:image_patches_three_iter}
\end{figure}

Figure \ref{fig:image_patches_three_iter} shows how our method gradually refines the pose for a few example frames of the \lmo{} dataset, illustrated by the image patches of a few iterations.
Despite the sub-optimal pose proposals from PVNet \cite{Peng_2019_CVPR}, the poses are accurately recovered.

\subsection{Detailed Pose Refinement Results}\label{sec:detailed_refinement_results}
Here we present detailed (per-object) pose refinement results and corresponding comparison with other methods.

Tables \ref{tab:real_lmo_add}, \ref{tab:real_lmo_reproj} and \ref{tab:real_lmo_degcm} show results on \lmo{} for the \add{}, \reprojs{} and \degcms{} metrics, respectively. The results of the corresponding experiments on synthetic data are reported in Tables \ref{tab:synth_lmo_add}, \ref{tab:synth_lmo_reproj} and \ref{tab:synth_lmo_degcm}.

Similarly, results on \lm{} are reported in Tables \ref{tab:real_lm_add}, \ref{tab:real_lm_reproj} and \ref{tab:real_lm_degcm}, for the \add{}, \reproj{} and \degcm{} metrics, respectively.

The symmetric objects \texttt{eggbox} and \texttt{glue} are marked with $^*$, and for them \add{} refers to \addsymm{}, and the \reprojs{} and \degcms{} metrics also take their ambiguities through $180$ degree rotations around the "up"-axis into account.

\begin{table}
    \centering
    \begin{tabular}{|r|c|c|c|c|}
        \cline{2-5}
        \multicolumn{1}{c|}{} &
            \thead{Oberweger \etal{} \cite{Oberweger_2018_ECCV}} &
            \thead{PVNet \cite{Peng_2019_CVPR}} &
            \thead{PoseCNN \cite{xiang2017posecnn}\\ + DeepIM \cite{li2019ijcv}} &
            \thead{PVNet \cite{Peng_2019_CVPR}\\ + PPC (Ours)} \\
        \hline
        ape &
            \score{17,60} &
            \score{15,04} &
            \scorebf{59,18} &
            \score{40,85} \\
        \hline
        can &
            \score{53,90} &
            \score{63,21} &
            \score{63,52} &
            \scorebf{82,44} \\
        \hline
        cat &
            \phantom{$0$}\score{3,31} &
            \score{20,30} &
            \score{26,24} &
            \scorebf{35,64} \\
        \hline
        driller &
            \score{62,40} &
            \score{64,00} &
            \score{55,58} &
            \scorebf{71,33} \\
        \hline
        duck &
            \score{19,20} &
            \score{33,86} &
            \scorebf{52,41} &
            \score{49,08} \\
        \hline
        eggbox$^*$ &
            \score{25,90} &
            \score{43,32} &
            \scorebf{62,95} &
            \score{57,28} \\
        \hline
        glue$^*$ &
            \score{39,60} &
            \score{49,83} &
            \scorebf{71,66} &
            \score{62,90} \\
        \hline
        holepuncher &
            \score{21,30} &
            \score{41,40} &
            \scorebf{52,48} &
            \score{43,14} \\
        \hline
        \textbf{Mean} &
            \score{30,40} &
            \score{41,37} &
            \scorebf{55,50} &
            \score{55,33}\\
        \hline
    \end{tabular}
    \caption{Results on \lmo{} according to the \add{} metric.}
    \label{tab:real_lmo_add}
\end{table}
\begin{table}
    \centering
    \begin{tabular}{|r|c|c|c|c|}
        \cline{2-5}
        \multicolumn{1}{c|}{} &
            \thead{Oberweger \etal{} \cite{Oberweger_2018_ECCV}} &
            \thead{PVNet \cite{Peng_2019_CVPR}} &
            \thead{PoseCNN \cite{xiang2017posecnn}\\ + DeepIM \cite{li2019ijcv}} &
            \thead{PVNet \cite{Peng_2019_CVPR}\\ + PPC (Ours)} \\
        \hline
        ape &
            \scorebf{69,60} &
            \score{66,84} &
            \score{69,02} &
            \score{68,97} \\
        \hline
        can &
            \score{82,60} &
            \scorebf{82,85} &
            \score{56,14} &
            \score{79,29} \\
        \hline
        cat &
            \score{65,10} &
            \score{62,34} &
            \score{50,95} &
            \scorebf{66,47} \\
        \hline
        driller &
            \score{73,80} &
            \score{70,68} &
            \score{52,94} &
            \scorebf{76,52} \\
        \hline
        duck &
            \score{61,40} &
            \score{59,58} &
            \score{60,54} &
            \scorebf{66,93} \\
        \hline
        eggbox$^*$ &
            \score{13,10} &
            \score{34,55} &
            \score{49,18} &
            \scorebf{49,28} \\
        \hline
        glue$^*$ &
            \scorebf{54,90} &
            \score{47,72} &
            \score{52,92} &
            \score{48,06} \\
        \hline
        holepuncher &
            \score{66,40} &
            \score{70,17} &
            \score{61,16} &
            \scorebf{75,45} \\
        \hline
        \textbf{Mean} &
            \score{60,86} &
            \score{61,84} &
            \score{56,61} &
            \scorebf{66,37}\\
        \hline
    \end{tabular}
    \caption{Results on \lmo{} according to the \reprojs{} metric. Note that \cite{Oberweger_2018_ECCV} reports results according to \reproj{}.}
    \label{tab:real_lmo_reproj}
\end{table}

\begin{table}
    \centering
    \begin{tabular}{|r|c|c|c|c|}
        \cline{2-4}
        \multicolumn{1}{c|}{} &
            \thead{PVNet \cite{Peng_2019_CVPR}} &
            \thead{PoseCNN \cite{xiang2017posecnn}\\ + DeepIM \cite{li2019ijcv}} &
            \thead{PVNet \cite{Peng_2019_CVPR}\\ + PPC (Ours)} \\
        \hline
        ape &
            \score{37,18} &
            \scorebf{51,75} &
            \score{47,69} \\
        \hline
        can &
            \score{63,38} &
            \score{35,82} &
            \scorebf{63,63} \\
        \hline
        cat &
            \score{19,43} &
            \score{12,75} &
            \scorebf{33,19} \\
        \hline
        driller &
            \score{60,21} &
            \score{45,24} &
            \scorebf{67,46} \\
        \hline
        duck &
            \score{15,31} &
            \score{22,48} &
            \scorebf{23,01} \\
        \hline
        eggbox$^*$ &
            \score{10,47} &
            \score{17,81} &
            \scorebf{33,87} \\
        \hline
        glue$^*$ &
            \score{20,93} &
            \scorebf{42,73} &
            \score{25,80} \\
        \hline
        holepuncher &
            \scorebf{40,00} &
            \score{18,84} &
            \score{37,52} \\
        \hline
        \textbf{Mean} &
            \score{33,36} &
            \score{30,93} &
            \scorebf{41,52}\\
        \hline
    \end{tabular}
    \caption{Results on \lmo{} according to the \degcms{} metric. No results are reported by Oberweger \etal{} \cite{Oberweger_2018_ECCV} on this metric.}
    \label{tab:real_lmo_degcm}
\end{table}

\begin{table}
    \centering
    \begin{tabular}{|r|c|c|c|c|}
        \cline{2-3}
        \multicolumn{1}{c|}{} &
            \thead{CDPN-synth \cite{Li_2019_ICCV}} &
            \thead{CDPN-synth \cite{Li_2019_ICCV}\\ + PPC-synth (Ours)} \\
        \hline
        ape &
            \score{17,65} &
            \scorebf{29,95} \\
        \hline
        can &
            \score{13,57} &
            \scorebf{36,68} \\
        \hline
        cat &
            \score{14,29} &
            \scorebf{16,84} \\
        \hline
        driller &
            \phantom{$0$}\score{5,00} &
            \scorebf{12,50} \\
        \hline
        duck &
            \scorebf{20,74} &
            \score{20,21} \\
        \hline
        eggbox$^*$ &
            \scorebf{33,16} &
            \scorebf{33,16} \\
        \hline
        glue$^*$ &
            \score{26,62} &
            \scorebf{29,87} \\
        \hline
        holepuncher &
            \scorebf{24,00} &
            \phantom{$0$}\score{9,50} \\
        \hline
        \textbf{Mean} &
            \score{18,76} &
            \scorebf{23,59} \\
        \hline
    \end{tabular}
    \caption{Synthetic results on \lmo{} according to the \add{} metric.}
    \label{tab:synth_lmo_add}
\end{table}

\begin{table}
    \centering
    \begin{tabular}{|r|c|c|c|c|}
        \cline{2-3}
        \multicolumn{1}{c|}{} &
            \thead{CDPN-synth \cite{Li_2019_ICCV}} &
            \thead{CDPN-synth \cite{Li_2019_ICCV}\\ + PPC-synth (Ours)} \\
        \hline
        ape &
            \score{48,66} &
            \scorebf{59,89} \\
        \hline
        can &
            \score{24,62} &
            \scorebf{34,17} \\
        \hline
        cat &
            \score{35,20} &
            \scorebf{40,82} \\
        \hline
        driller &
            \phantom{$0$}\score{7,50} &
            \scorebf{15,00} \\
        \hline
        duck &
            \score{51,60} &
            \scorebf{54,79} \\
        \hline
        eggbox$^*$ &
            \score{34,20} &
            \scorebf{34,72} \\
        \hline
        glue$^*$ &
            \scorebf{14,94} &
            \score{12,99} \\
        \hline
        holepuncher &
            \scorebf{48,50} &
            \score{35,50} \\
        \hline
        \textbf{Mean} &
            \score{32,22} &
            \scorebf{35,99} \\
        \hline
    \end{tabular}
    \caption{Synthetic results on \lmo{} according to the \reprojs{} metric.}
    \label{tab:synth_lmo_reproj}
\end{table}

\begin{table}
    \centering
    \begin{tabular}{|r|c|c|c|c|}
        \cline{2-3}
        \multicolumn{1}{c|}{} &
            \thead{CDPN-synth \cite{Li_2019_ICCV}} &
            \thead{CDPN-synth \cite{Li_2019_ICCV}\\ + PPC-synth (Ours)} \\
        \hline
        ape &
            \score{26,74} &
            \scorebf{36,90} \\
        \hline
        can &
            \score{17,59} &
            \scorebf{26,13} \\
        \hline
        cat &
            \score{13,78} &
            \scorebf{18,88} \\
        \hline
        driller &
            \phantom{$0$}\score{6,50} &
            \scorebf{11,00} \\
        \hline
        duck &
            \score{15,43} &
            \scorebf{16,49} \\
        \hline
        eggbox$^*$ &
            \scorebf{30,57} &
            \score{23,83} \\
        \hline
        glue$^*$ &
            \phantom{$0$}\score{5,84} &
            \phantom{$0$}\scorebf{9,74} \\
        \hline
        holepuncher &
            \scorebf{19,00} &
            \score{15,50} \\
        \hline
        \textbf{Mean} &
            \score{16,13} &
            \scorebf{19,81} \\
        \hline
    \end{tabular}
    \caption{Synthetic results on \lmo{} according to the \degcms{} metric.}
    \label{tab:synth_lmo_degcm}
\end{table}

\begin{table}
    \centering
    \begin{tabular}{|r|c|c|c|c|}
        \cline{2-4}
        \multicolumn{1}{c|}{} &
            \thead{PoseCNN \cite{xiang2017posecnn}} &
            \thead{PoseCNN \cite{xiang2017posecnn}\\ + DeepIM \cite{li2019ijcv}} &
            \thead{PoseCNN  \cite{xiang2017posecnn}\\ + PPC (Ours)} \\
        \hline
        ape &
            \score{27,71} &
            \scorebf{76,95} &
            \score{75,14} \\
        \hline
        benchvise &
            \score{68,87} &
            \scorebf{97,48} &
            \score{94,28} \\
        \hline
        camera &
            \score{47,35} &
            \score{93,53} &
            \scorebf{96,18} \\
        \hline
        can &
            \score{71,33} &
            \score{92,81} &
            \scorebf{96,95} \\
        \hline
        cat &
            \score{56,64} &
            \score{82,14} &
            \scorebf{89,82} \\
        \hline
        driller &
            \score{65,28} &
            \score{94,95} &
            \scorebf{97,92} \\
        \hline
        duck &
            \score{42,86} &
            \scorebf{77,65} &
            \score{69,39} \\
        \hline
        eggbox$^*$ &
            \score{97,84} &
            \score{97,09} &
            \scorebf{98,59} \\
        \hline
        glue$^*$ &
            \score{94,88} &
            \scorebf{99,42} &
            \score{92,95} \\
        \hline
        holepuncher &
            \score{44,00} &
            \score{52,81} &
            \scorebf{68,70} \\
        \hline
        iron &
            \score{65,47} &
            \scorebf{98,26} &
            \score{90,19} \\
        \hline
        lamp &
            \score{69,96} &
            \score{97,50} &
            \scorebf{98,27} \\
        \hline
        phone &
            \score{54,39} &
            \scorebf{87,72} &
            \score{84,32} \\
        \hline
        \textbf{Mean} &
            \score{62,04} &
            \score{88,33} &
            \scorebf{88,67}\\
        \hline
    \end{tabular}
    \caption{Results on \lm{} according to the \add{} metric.}
    \label{tab:real_lm_add}
\end{table}

\begin{table}
    \centering
    \begin{tabular}{|r|c|c|c|c|}
        \cline{2-4}
        \multicolumn{1}{c|}{} &
            \thead{PoseCNN \cite{xiang2017posecnn}} &
            \thead{PoseCNN \cite{xiang2017posecnn}\\ + DeepIM \cite{li2019ijcv}} &
            \thead{PoseCNN  \cite{xiang2017posecnn}\\ + PPC (Ours)} \\
        \hline
        ape &
            \score{82,67} &
            \scorebf{98,38} &
            \score{97,71} \\
        \hline
        benchvise &
            \score{49,95} &
            \score{96,99} &
            \scorebf{97,87} \\
        \hline
        camera &
            \score{71,67} &
            \scorebf{98,92} &
            \score{98,63} \\
        \hline
        can &
            \score{69,85} &
            \scorebf{99,70} &
            \score{97,64} \\
        \hline
        cat &
            \score{92,01} &
            \score{98,70} &
            \scorebf{98,80} \\
        \hline
        driller &
            \score{43,45} &
            \score{96,13} &
            \scorebf{97,13} \\
        \hline
        duck &
            \score{91,73} &
            \scorebf{98,5} &
            \score{97,93} \\
        \hline
        eggbox$^*$ &
            \score{41,82} &
            \score{96,15} &
            \scorebf{98,50} \\
        \hline
        glue$^*$ &
            \score{87,73} &
            \scorebf{98,94} &
            \score{96,24} \\
        \hline
        holepuncher &
            \score{59,52} &
            \score{96,29} &
            \scorebf{98,57} \\
        \hline
        iron &
            \score{41,68} &
            \scorebf{97,24} &
            \scorebf{97,24} \\
        \hline
        lamp &
            \score{48,27} &
            \scorebf{94,24} &
            \score{94,15} \\
        \hline
        phone &
            \score{58,46} &
            \score{97,73} &
            \scorebf{98,39} \\
        \hline
        \textbf{Mean} &
            \score{64,52} &
            \score{97,53} &
            \scorebf{97,60}\\
        \hline
    \end{tabular}
    \caption{Results on \lm{} according to the \reproj{} metric. Note that \cite{li2019ijcv} reports results according to \reprojs{}.}
    \label{tab:real_lm_reproj}
\end{table}

\begin{table}
    \centering
    \begin{tabular}{|r|c|c|c|}
        \cline{2-4}
        \multicolumn{1}{c|}{} &
            \thead{PoseCNN \cite{xiang2017posecnn}} &
            \thead{PoseCNN \cite{xiang2017posecnn}\\ + DeepIM \cite{li2019ijcv}} &
            \thead{PoseCNN  \cite{xiang2017posecnn}\\ + PPC (Ours)} \\
        \hline
        ape &
            \phantom{$0$}\score{6,95} &
            \score{90,38} &
            \scorebf{96,48} \\
        \hline
        benchvise &
            \score{13,58} &
            \score{88,65} &
            \scorebf{90,40} \\
        \hline
        camera &
            \score{20,39} &
            \scorebf{95,78} &
            \score{91,67} \\
        \hline
        can &
            \score{24,39} &
            \score{92,81} &
            \scorebf{94,39} \\
        \hline
        cat &
            \score{24,98} &
            \score{87,62} &
            \scorebf{96,01} \\
        \hline
        driller &
            \score{18,25} &
            \score{92,86} &
            \scorebf{96,13} \\
        \hline
        duck &
            \score{18,23} &
            \scorebf{85,16} &
            \score{83,10} \\
        \hline
        eggbox$^*$ &
            \score{16,53} &
            \score{63,85} &
            \scorebf{95,49} \\
        \hline
        glue$^*$ &
            \score{19,50} &
            \scorebf{83,01} &
            \score{73,07} \\
        \hline
        holepuncher &
            \score{15,81} &
            \score{54,52} &
            \scorebf{82,11} \\
        \hline
        iron &
            \score{12,97} &
            \scorebf{92,65} &
            \score{92,03} \\
        \hline
        lamp &
            \score{24,38} &
            \score{90,88} &
            \scorebf{92,51} \\
        \hline
        phone &
            \score{19,26} &
            \scorebf{89,16} &
            \score{83,29} \\
        \hline
        \textbf{Mean} &
            \score{18,14} &
            \score{85,21} &
            \scorebf{89,74}\\
        \hline
    \end{tabular}
    \caption{Results on \lm{} according to the \degcm{} metric. Note that \cite{li2019ijcv} reports results according to \degcms{}.}
    \label{tab:real_lm_degcm}
\end{table}

\section{Additional Notes}

\subsection{Negative Depth Correction of Pose Proposals}\label{sec:neg_label_correction}
We observed that the pose proposals from PVNet \cite{Peng_2019_CVPR} sometimes have negative depth, and in this case we switched sign for the object center position, and rotated the object 180 degrees around the principal axis of the camera, in order to yield a feasible estimate with similar projection (the projection is identical for points on the plane which goes through the object center and is parallel to the principal plane of the camera).
This correction is done both when reporting the results of \cite{Peng_2019_CVPR}, and when reporting the results of our refinement.

\clearpage
\bibliographystyle{unsrt}  
\bibliography{ppc}

\end{document}